\documentclass[acmsmall]{acmart}

\AtBeginDocument{%
  }

\usepackage{svg}
\usepackage{xcolor}
\usepackage{pifont}
\usepackage{adjustbox}
\usepackage{multirow}

\newcommand{\vcheck}[0]{\textcolor{green}{\ding{51}}}
\newcommand{\xmark}[0]{\textcolor{red}{\ding{55}}}

\begin{document}

\title{MARS: Paying more attention to visual attributes for text-based person search}

\author{Alex Ergasti}
\email{alex.ergasti@unipr.it}
\authornote{Authors contributed equally to this work}
\author{Tomaso Fontanini}
\authornotemark[1]
\email{tomaso.fontanini@unipr.it}
\affiliation{%
  \institution{Department of Engineering and Architecture, University of Parma}
  \city{Parma}
  \state{Italy}
  \country{IT}
}

\author{Claudio Ferrari}
\affiliation{%
  \institution{Department of Engineering and Architecture, University of Parma}
  \city{Parma}
  \state{Italy}
  \country{IT}
}
\email{claudio.ferrari2@unipr.it}

\author{Massimo Bertozzi}
\affiliation{%
  \institution{Department of Engineering and Architecture, University of Parma}
  \city{Parma}
  \state{Italy}
  \country{IT}
}
 \email{massimo.bertozzi@unipr.it}

\author{Andrea Prati}
\affiliation{%
  \institution{Department of Engineering and Architecture, University of Parma}
  \city{Parma}
  \state{Italy}
  \country{IT}
}
  \email{andrea.prati@unipr.it}

\renewcommand{\shortauthors}{Ergasti et al.}

\begin{abstract}

Text-based person search (TBPS) is a problem that gained significant interest within the research community. The task is that of retrieving one or more images of a specific individual based on a textual description. The multi-modal nature of the task requires learning representations that bridge text and image data within a shared latent space. Existing TBPS systems face two major challenges. One is defined as \textit{inter-identity} noise that is due to the inherent vagueness and imprecision of text descriptions and it indicates how descriptions of visual attributes can be generally associated to different people; the other is the \textit{intra-identity} variations, which are all those nuisances \textit{e.g.} pose, illumination, that can alter the visual appearance of the same textual attributes for a given subject. To address these issues, this paper presents a novel TBPS architecture named MARS (Mae-Attribute-Relation-Sensitive), which enhances current state-of-the-art models by introducing two key components: a Visual Reconstruction Loss and an Attribute Loss. The former employs a Masked AutoEncoder trained to reconstruct randomly masked image patches with the aid of the textual description. In doing so the model is encouraged to learn more expressive representations and textual-visual relations in the latent space. 
The Attribute Loss, instead, balances the contribution of different types of attributes, defined as adjective-noun chunks of text. This loss ensures that every attribute is taken into consideration in the person retrieval process. 
Extensive experiments on three commonly used datasets, namely CUHK-PEDES, ICFG-PEDES, and RSTPReid, report performance improvements, with significant gains in the mean Average Precision (mAP) metric w.r.t. the current state of the art. Code will be available at \href{https://github.com/ErgastiAlex/MARS}{https://github.com/ErgastiAlex/MARS}.
\end{abstract}



\received{30 June 2024}
\received[revised]{---}
\received[accepted]{---}

\maketitle

\section{Introduction}\label{sec:introduction}
\begin{figure}[b!]
    \centering
    \includegraphics[width=0.9\textwidth]{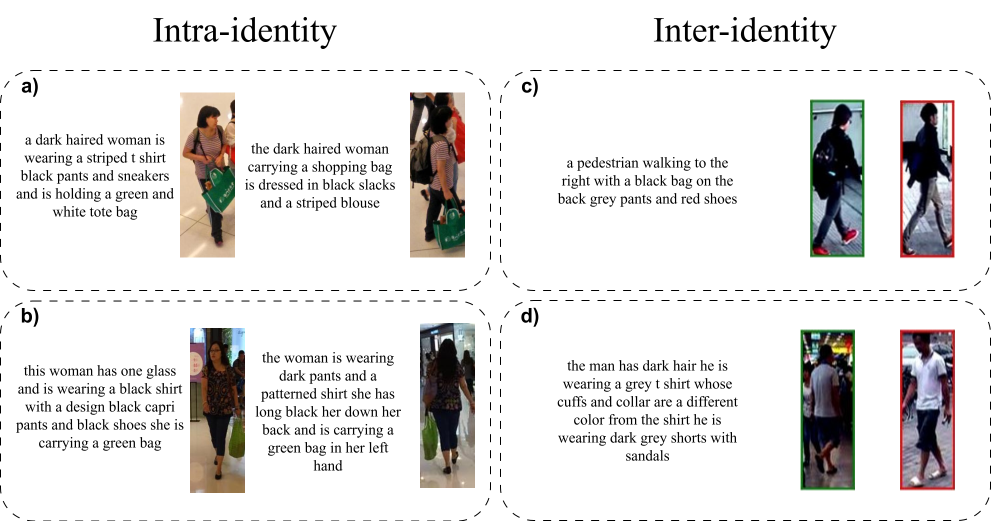}
    \caption{CUHK-PEDES images and caption. On the left, \textit{a} and \textit{b} are examples of intra-identity variations where the visual attributes of the same person (\textit{e.g.}, pose, illumination, etc..) vary between images. On the right,  \textit{c} and \textit{d} are examples of inter-identity variations where a caption can be matched to two identities which look very similar between each others but only one is correct (green for correct match, red for wrong match).}
    \label{fig:intra-inter}
\end{figure}
The integration of text prompts in the re-identification task, called text-based person search (TBPS), has gained lots of interest in the research community lately \cite{bai2023rasa, niu2024comprehensive}. In TBPS, textual descriptions are queries used to search \textit{specific identities} in a gallery of images. This is similar yet conceptually different from the standard text-based image retrieval task, in which captions are used to find one or multiple images that best match the given description. Commonly, architectures designed for TBPS include two encoders, one for images and one for the text prompts. The encoders extract a latent code for each modality which can be then aligned using various loss functions such as cross-modal projection matching \cite{zhang2018deep} or contrastive loss \cite{bai2023text}. By doing so, textual and visual latent codes are forced to lie in a common space, so that one can use the text embeddings to retrieve the latent code of the most similar image. The most popular choice opted by recent approaches, \textit{e.g.} \cite{jiang2023cross, bai2023rasa, lin2024cross}, is to fine-tune and adapt pre-trained large vision-language models such as CLIP \cite{yan2023clip}, BLIP \cite{li2022blip} and ALBEF \cite{li2021align}. This is motivated by the relative small size of datasets commonly used in TBPS, which are typically composed by less than 100k images. The fine grained knowledge that is provided by such large models can be used as a solid starting point to train a TBPS system. Additionally, architectures based on BLIP~\cite{lin2024cross} or ALBEF~\cite{bai2023rasa} use a cross-modal encoder that fuses together image and text information via cross-attentions layers and performs an additional matching. More in detail, in such architectures, the searching task is composed of two phases: in the first phase, for each textual embedding, a list of $k$ nearest-neighbor images is obtained; then, a re-ranking of the top $k$ candidates is performed based on the matching results of the cross-modal encoder. 

Using text in place of images to perform retrieval opens up both several new possibilities and new challenges. On the one hand, a query image is no longer required, resulting in a more flexible and easy search procedure. On the other hand, text prompts are often vague or ambiguous, and lack the objectivity that images instead can provide. Captions included in standard datasets like ``A girl with a black bag and white shirt'' lack the necessary unique details that are needed to distinguish similar images. For example, the bag could be both on the right or on the left shoulder, or the shirt could have different details such as logos or textures. Such differences are slight, yet they might correspond to different identities in a given video. This vagueness hinders the quantitative results in a TBPS system in which we care about finding precise identities given the captions and not just the most similar images. We refer to this as \textit{inter-identity} noise (see Fig. \ref{fig:intra-inter} on the right).

Another key problem is represented by the \textit{intra-identity} variations (see Fig. \ref{fig:intra-inter} on the left). The appearance of the same subject in the dataset can vary depending on several factors such as pose, camera position (front or back facing), or illumination. At the same time, different text descriptions can be used to describe the subject with different level of granularity and ambiguity. These nuisances might have a non-negligible effect; for example, if a person is captured from the back, attributes such as ``man/woman'' become even more ambiguous. 

Several approaches proposed solutions to limit this problem. The most common one consists in building a more fine-grained relationship between image and text embeddings by performing masked language modeling \cite{jiang2023cross, lin2024cross}. This is achieved by masking the text prompt and, via cross-attention mechanism, utilizing the image patch embeddings to predict the missing words. Alternatively, RaSa \cite{bai2023rasa} proposed a slightly different solution in addition to masking, which consists in changing some words, and then training the model to recognize which words were changed.
In addition to the above, in this work we argue that another problem of current TBPS systems is that the existing text encoding techniques do not fully exploit all the attributes contained in a given text, making the retrieval less precise. Indeed, assigning the same importance to all attributes, especially to the most discriminative ones, is often fundamental to distinguish different identities. In fact, two different subjects, that are yet very similar in appearance, might only be correctly separated by a single small attribute \textit{e.g.} shoes color, in their description. This is true in particular for long textual description containing several attributes, where we want the TBPS system to balance the contribution of each attribute equally during the retrieval.  
The attribute loss proposed in this work was designed precisely to push the model to correctly exploit all the attributes.

In this paper, we present a novel TBPS architecture named MARS (Mae-Attribute-Relation-Sensitive) that attempts to further improve current state-of-the-art architectures. The proposed system is composed by a text encoder, an image encoder, a cross-modal encoder and a masked autoencoder. Additionally, it introduces two novel losses during training.

Firstly, a novel attribute loss is proposed that matches each set of attributes in the captions and the image data. This pushes the cross-modal encoder to consider each attribute in a caption with equal weight and, as a consequence, reduces the uncertainty in the retrieval. Differently from other approaches such as \cite{niu2024comprehensive}, where every word except adverbs, determiners, special characters and numerals is considered an ``attribute'', we define the attributes in a sentence as a set of words following the structure adjective+noun (\textit{e.g.} ``white shirt''). Each of these set forms an \textit{attribute chunk}. The matching is performed in the output of the cross modal encoders, where the average of the embeddings corresponding to each attribute chunk is classified strengthening the correlation between textual and image data.

Secondly, to further enhance the capability of the text and image encoder, we add a loss inspired by the Masked AutoEncoder (MAE) architecture \cite{he2022masked}. In MAE, the input image of the encoder is masked (\textit{i.e.}, some patches are removed) and the decoder is tasked to reconstruct the original images. Specifically, in MARS the image encoder acts as the MAE encoder and an additional MAE decoder is added to perform reconstruction. Furthermore, the decoder takes as input also the embeddings extracted from the text to help guide the image reconstruction. In this way, we aim to further enhance the mutual-information encapsulated in both image and text embedding.

Finally, the key contributions of this work are the following:
\begin{itemize}
    \item \textbf{MARS}: a novel TBPS architecture is proposed which is composed by four main components: a text encoder and an image encoder that embed text descriptions and images, a cross-modal encoder with additional cross attention layers w.r.t. the current state of the art that fuses textual and image embeddings to perform an additional matching and finally, a novel masked autoencoder that performs reconstruction over masked image patches with the help of textual information.
    \item \textbf{Attribute Loss}: We present a novel attribute loss, which aims at improving the matching accuracy between text and image at the attribute level. This loss matches each set of attributes in a given text with the image.  This approach enhances the model capability to provide to each attribute in a given text descriptions equal weight, in order to accurately discriminate between different identities. By doing so, the attribute loss allows the model to put attention on both common and rare attributes in the retrieval process.
    \item \textbf{Masked AutoEncoder Loss}: We present a Masked AutoEncoder loss which aims to reinforce the mutual-information encapsulated in each embedding. This method uses the Image Encoder as a MAE encoder and adds a new light weight decoder which also takes as input the text embedding in order to reconstruct the original image.
\end{itemize}

\section{Related Works}
Joining together text and images for the task of text-based image retrieval and tracking was first explored by Shuang, \textit{et al.} \cite{li2017person}, who also introduced the CUHK-PEDES dataset. This dataset is composed of a set of pedestrian images paired with a text description which serves as query to retrieve the correct subject. This new dataset and problem to be solved garnered a lot of attention, and several methods were proposed to address it. Zheng \textit{et al.} \cite{zheng2020hierarchical} proposed a novel hierarchical Gumbel attention network to boost cross-modal alignment, while Wang \textit{et al.} \cite{wang2021text} introduced a novel multi-granularity embedding learning model. On the other side, \cite{zhang2018deep} proposed a cross-modal projection matching (CMPM) loss and a cross-modal projection classification (CMPC) loss. Later, Shao \textit{et al.} \cite{shao2022learning} introduced an end-to-end framework based on transformers to learn, for both text and images, granularity-unified representations. In addition, a set of methods experimented with using additional data such as segmentation, pose estimation or attribute prediction to boost the retrieval performance \cite{wang2020vitaa,zhu2021dssl}.

In addition, Wu \textit{et al.} \cite{wu2021lapscore} introduced two sub-tasks, image colorization and text completion. The first one helps learning rich text information to colorize gray images, while, in the second one, the model is requested to complete color word vacancies in the captions. Then, Zeng \textit{et al.} \cite{zeng2022relation} proposed a Relation-aware Aggregation Network (RAN) exploiting the relationship between the person and the local objects. Additionally, three auxiliary tasks are introduced: identifying the gender of the pedestrian, discerning the images of the similar pedestrian, and aligning the semantic information between caption and image. Also, a common problem in text-to-image search is the presence of weak positive pairs. This was first tackled by Ding \textit{et al.} \cite{ding2021semantically} that assigned different margins in the triplet loss.

Up until this point, the vision encoder and the text encoder necessary to align the embeddings of the different modalities were trained from scratch. Recently, the use of pretrained vision-language models has caught attention, \textit{e.g.} in \cite{shu2022see, yan2023clip, cao2024empirical, yan2023CLIPdrive}. Cao \textit{et al.} \cite{yan2023clip} perform an empirical study about using CLIP \cite{radford2021learning} as backbone for TBPS. Among these, IRRA \cite{jiang2023cross}, which was pretrained on CLIP, introduced an Implicit Relation Reasoning module and aims to minimize the KL divergence between distributions of image-text similarity and normalized label matching. Also, IRRA proposed a masked language modelling (MLM) in which a masked set of image embeddings is reconstructed with the aid of text tokens.  Additionally, RaSa \cite{bai2023rasa} designed two novel strategies: Relation-Aware learning (RA) and Sensitivity-Aware learning (SA). A concurrent work with RaSa is represented by CADA \cite{lin2024cross} which focuses of building bidirectional image-text associations. More in detail, it tries to associate text tokens with image patches and image regions with text attributes. The latter is done by modifying the MLM into masking specific attributes and not random words.

In addition to pretraining on common text-image datasets not specifically tailored to pedestrian identification, Yang \textit{et al.} \cite{yang2023towards} introduced a novel dataset named MALS (Multi-Attribute and Language Search). 
The MALS dataset was generated using diffusion models to overcome privacy concerns and annotation costs associated with real-world data collection. To evaluate the effectiveness of this dataset, Yang \textit{et al.} developed a model called APTM (Attribute Prompt Learning and Text Matching Learning).



In APTM the authors proposed a new attribute loss, named Image-Attribute Matching (IAM) loss. This loss function is designed to classify image-text pairs $(I,T)$ using concise text descriptions $T$ that contain only partial information about the subject (\textit{e.g.}, "The person wears pants or shorts"). On the contrary, in our paper, we propose a structured Attribute Loss with the purpose of pushing the cross-modal encoder to perform the match between image and text using each of the attributes contained in the captions. In particular, our loss does not build a new caption as in \cite{yang2023towards}, but pushes the model to focus more on the attribute embeddings in an explicit manner. More in details, our model performs an additional matching between image and text based on each of the attributes contained in the sentence.




\section{Proposed Method}
In this section, the proposed model architecture will be presented as well as the training losses.

\begin{figure}
    \includegraphics[width=\textwidth]{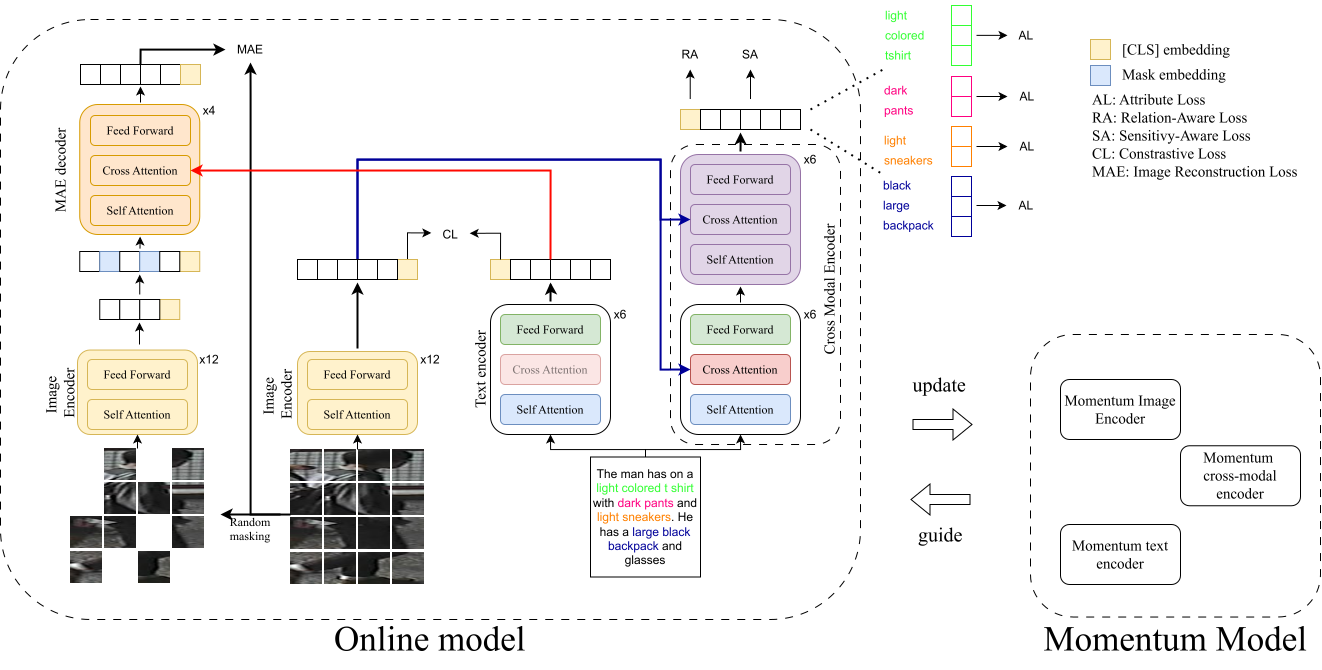}
    \caption{Overview of the proposed architecture (same color corresponds to shared parameters). Firstly, an input pair of image and text $(I,T)$ is fed to the Image Encoder $\mathcal{E}_v$ and the Text Encoder $\mathcal{E}_t$, respectively, and Contrastive Loss is applied to the obtained embeddings $\mathbf{v}$ and $\mathbf{t}$. Secondly, the MAE Decoder $\mathcal{D}_{mae}$ is trained to reconstruct a masked image patches sequence into the original unmasked one. Finally, text is fed to the Cross-Modal Encoder $\mathcal{E}_{cross}$ and the visual embeddings $\mathbf{v}$ are injected into its cross-attention layers. The output of $\mathcal{E}_{cross}$ $\mathbf{f}$ is employed into three different loss functions: (a) the class token $f_{cls}$ is used in the Relation-Aware Loss to learn a matching function between positive and negative image-text pairs, then, (b) given a masked input text $T_{mask}$ Sensitive-Aware Loss is used to identify the masked word and finally, (c) the Attribute Loss is calculated over the embeddings corresponding to attributes chunks in the text.} 
    \label{fig:architecture}
\end{figure}

\subsection{The MARS Architecture}
In this paper we propose MARS (Mae-Attribute-Relation-Sensitive), a novel architecture for TBPS. When building the system, we decided to use RaSa \cite{bai2023rasa} as starting point since currently is one of the best TBPS models and we initialized the architecture weights on ALBEF \cite{li2021align}. 

MARS is composed by four main components (Fig. \ref{fig:architecture}): (a) an Image Encoder $\mathcal{E}_v$ which encodes a sequence of image patches, (b) a Text Encoder $\mathcal{E}_t$ which produces the text embeddings from the captions, (c) a MAE Decoder $\mathcal{D}_{mae}$ which is tasked to reconstruct masked images and, finally, (d) a Cross-Modal Encoder $\mathcal{E}_{cross}$ which computes our proposed attribute loss along with the baseline RaSa \cite{bai2023rasa} losses: Sensitive-Aware and Relation-Aware losses. 

More in detail, the Image Encoder is a Vision Transformer (ViT)~\cite{dosovitskiy2020image} composed by 12 transformer blocks consisting in Self-Attention layers and Feed Forward Layers. The Text Encoder and the Cross-Modal Encoder are based on BERT \cite{devlin2018bert} which is a 12 blocks transformer-based architecture for language understanding. The first 6 blocks of BERT are used as Text Encoder. On the other hand, the Cross-Modal Encoder is composed by all the 12 blocks of BERT, but, differently than previous methods like \cite{bai2023rasa, li2021align}, we equip all its blocks with cross-attention layers instead of only the last 6. By doing so, we can perform the cross-modal encoding using the whole BERT architecture, which helps boosting the matching accuracy as it will be shown in the experiments. Finally, the MAE Decoder is composed by 4 transformer blocks equipped with cross attentions. Additionally, a momentum model is initialized. The momentum model is a slower version of the online model whose weights are obtained using Exponential Moving Average (EMA):
\begin{equation}
    \hat{\theta} = m\hat{\theta} + (1-m)\theta
\end{equation}
\noindent where $\hat{\theta}$ are the weights of the momentum models, while $\theta$ are the weights of the online model, and $m$ is the momentum coefficient. This model will be crucial when calculating the losses as explained in Section \ref{sec:losses}. 

During training, starting from a image-text pair $(I,T)$, $\mathcal{E}_v$ produces a sequence of image embeddings $\mathbf{v} = \{v_{cls}, v_1, \cdots, v_M\}$ for each of the $M$ image patches, while a tokenized text is fed to $\mathcal{E}_t$ producing a sequence of text embeddings $\mathbf{t} = \{t_{cls}, t_1, \cdots, t_N\}$, being $N$ the number of word. In both $\mathbf{v}$ and $\mathbf{t}$ the first embedding is the class token \texttt{[CLS]}. Additionally, a masked version of the image patches of length $L<M$ is embedded using $\mathcal{E}_v$. Then, a set of $K = M-L$ mask embeddings are inserted in the obtained sequence at the masked positions and the whole sequence is fed to $\mathcal{D}_{mae}$ which reconstructs the original image also with the aid of text embeddings $\mathbf{t}$ that are fed in $\mathcal{D}_{mae}$ via cross attention mechanism.
Finally, text $T$ is used as input to $\mathcal{E}_{cross}$ while image embeddings $\mathbf{v}$ are injected in $\mathcal{E}_{cross}$ cross attention layers producing the cross-modal embeddings $\mathbf{f} =  \{f_{cls}, f_1, \cdots, f_N\}$. The \texttt{[CLS]} token of the cross-modal embeddings will be used to perform an additional matching between images and captions.

The evaluation phase is composed of two steps: first, all the image and text embeddings are calculated using the image and text encoder and, for each text embedding, an ordered list of the closest image embedding is obtained by calculating the similarity between the \texttt{[CLS]} token of the text and the images. Then, the first $k$ candidates for each text are selected and an additional re-ranking phase is performed considering the matching results of the Cross-Modal Encoder. This additional step allows to further boost the ranking results.

\subsection{Baseline Losses}\label{sec:losses}
As a baseline training objective for our model, we employ the loss set used in RaSa \cite{bai2023rasa}. Additionaly, our final proposed architecture also introduces two novel losses: an Attribute Loss and a Masked Autoencoder Loss. 

\paragraph{Relation-Aware Loss.}
The Relation-Aware (RA) loss is a modification to the conventional Image-Text Matching (ITM) loss commonly employed in various models \cite{li2021align,li2022blip,yang2023towards}. In particular, ITM performs a binary classification between positive and negative image-text pairs. Instead of selecting hard-negative samples at random, the ITM variation, denoted as $p$-ITM, creates a negative pair set by evaluating embedding similarity and employing this value as the probability of drawing a negative pair. This similarity is quantified using the \texttt{[CLS]} token representations from the unimodal encoders (Text and Image Encoder in Fig.~\ref{fig:architecture}). The probability of choosing a negative pair is proportional to the similarity of the corresponding image-text \texttt{[CLS]} tokens. Consequently, negative pairs exhibiting higher similarity are more likely to be selected, enhancing the robustness of the model in distinguishing between truly-related and unrelated image-text pairs. The loss $\mathcal{L}_{p-ITM}$ is a Cross-Entropy Loss that distinguishes if input pairs $(I, T)$ are positive or negative.

Let $l^{itm}_c(\mathbf{f}_{cls})$ be a fully connected layer applied on the \texttt{[CLS]} token of $\mathcal{E}_{cross}(T,\mathcal{E}_v(I))$ which predicts the logit for a given class $c$. The loss can be calculated as:
\begin{equation}
    \mathcal{L}_{p-ITM}=-\frac{1}{3\cdot N_B}\sum_{(I, T)\in P}\sum_{c\in C}y_c\log\frac{\exp(l^{itm}_c(\mathbf{f}_{cls}))}{\sum_{n\in C}\exp(l^{itm}_n(\mathbf{f}_{cls}))}
    \label{eq:lpitm}
\end{equation}
where $C$ is the set of possible classes, which includes two categories: positive and negative pairs. The variable $y_c$ represents the ground-truth, where $y_c=1$ if the pair $(I, T)$ belongs to the class $c$. The set $P$ is built as the union of three subsets, hence the division by 3, each of size $N_B$, $P^{++},P^{-+},P^{+-}$:
\begin{itemize}
    \item \( P^{++} \) consists of the input batch, where all pairs \((I, T)\) are positive.
    \item \( P^{-+} \) is composed of a negative image \( I \) for each text \( T \), sampled randomly with a probability determined by the similarity between $t_{cls}$ obtained from $\mathcal{E}_{t}(T)$ and $v_{cls}$ obtained from $\mathcal{E}_v(I)$.
    \item \( P^{+-} \) is composed of a negative text \( T \) for each image \( I \), sampled randomly with a probability determined by the similarity between $v_{cls}$ obtained from $\mathcal{E}_{v}(I)$ and $t_{cls}$ obtained from $\mathcal{E}_t(T)$.
\end{itemize}

\noindent
Furthermore, the $p$-ITM loss is expanded by adding a Positive Relation Detection (PRD), formulated as a Cross Entropy Loss, which aims to detect weak positive pairs. During training, the weak positive pairs are built by randomly switching the caption of an image with a caption of a different image having the same identity. Viceversa, we define strong positive pairs as the original pairs coming from the dataset. Let $l^{prd}_c(f_{cls})$ be a fully connected layer applied on the \texttt{[CLS]} token of $\mathcal{E}_{cross}(T,\mathcal{E}_v(I))$ which predict the logit for a given class $c$, then:

\begin{equation}
        \mathcal{L}_{prd}=-\frac{1}{N_B}\sum_{(I, T)\in P^{++}}\sum_{c\in C}y_c\log\frac{\exp(l^{prd}_c(f_{cls}))}{\sum_{n\in C}\exp(l^{prd}_n(f_{cls}))}
\end{equation}
where $P^{++}$ are only positive pairs that can be both weak or strong and $C$ is the number of classes (two in this case), corresponding to strong positive pairs and weak positive pairs. The final RA loss is then computed as:
\begin{equation}
    \mathcal{L}_{RA}=\mathcal{L}_{p-ITM}+\lambda_1\mathcal{L}_{prd}
\end{equation}
\noindent where $\lambda_1$ is an hyperparameter used to balance the contribution of $\mathcal{L}_{prd}$.

\paragraph{Sensitive-Aware Loss.} 
Similar to RA loss, Sensitive-Aware (SA) loss is an expansion of the basic Masked Language Modeling (MLM) introduced in \cite{jiang2023cross} that adds a Momentum-based Replace Token Detection ($m$-RTD). Given a strongly positive pair \((I, T)\), the MLM loss is expressed as a Cross Entropy Loss. Given a masked text $T_{mask}$, where each word has a probability \( p \) of being masked out, the model is trained to predict the correct missing word. Let \( V \) represent the set of all possible words in the vocabulary and $l_v(\mathbf{f}_{mask})$ be a fully connected layer applied on each embedding obtained from $\mathcal{E}_{cross}(T_{mask},\mathcal{E}_v(I))$ which  predicts the logit for the vocabulary $v$. The MLM loss is formulated as:

\begin{equation}
    \mathcal{L}_{MLM}=-\frac{1}{N_B}\sum_{(I,T)\in P^{++}}\frac{1}{N_{mask}^t}\sum_{w\in t}m_w\sum_{v\in V}y_v\log\frac{exp(l_v(\mathbf{f}_{mask}))}{\sum_{n\in V}\exp(l_n(\mathbf{f}_{mask}))}
\end{equation}
where $N_B$ is the batch size, $N_{mask}^t$ is the number of masked words for a given text $t$, $m_w$ is 1 if the word is masked, otherwise 0 (\textit{i.e.} $N_{mask}^t=\sum_{w\in t}m_w$) and $y_v$ is a one-hot value on the ground-truth vocabulary. On the other hand, in $m$-RTD, the focus is on detecting words that have been replaced. To replace the masked word, the momentum model of the MLM is employed, which converges slowly providing less accurate word predictions. The MLM momentum model predicts a word for each masked word, by effectively replacing the masked words with its predictions, and the task of the online model is to identify which of these words have been replaced. The $m$-RTD loss is based on a Cross-Entropy Loss which teaches the model to distinguish between replaced and non-replaced words. Let $C$ be the set of possible predictions for each word, where a prediction can be either "replaced" or "not replaced", and $l_c(\mathbf{f}_{repl}))$ be a fully connected layer applied on each embedding obtained from $\mathcal{E}_{cross}(T_{repl},\mathcal{E}_v(I))$ which predicts the logit for the class $c$. The loss function can be expressed as:
\begin{equation}
    \mathcal{L}_{m-RTD}=-\frac{1}{N_B}\sum_{(I,T) \in P^{++}}\frac{1}{N_w^t}\sum_{c\in C}y_c\log \frac{\exp(l_c(\mathbf{f}_{repl})))}{\sum_{n\in C}\exp(l_n(\mathbf{f}_{repl})))}
\end{equation}
where $N_B$ is the batch size, $N_w^t$ is the number of words in a given text $t$ and $y_c$ is the ground-truth. The final $\mathcal{L}_{SA}$ is then:
\begin{equation}
    \mathcal{L}_{SA}=\mathcal{L}_{MLM}+\lambda_2\mathcal{L}_{m-RTD}
\end{equation}
\noindent where $\lambda_2$ is an hyperparameter used to balance the contribution of $\mathcal{L}_{m-RTD}$.

\paragraph{Contrastive Loss.}
Contrastive Loss (CL) is the last baseline model loss. As shown by Fig. ~\ref{fig:architecture}, the contrastive loss is calculated using only the \texttt{[CLS]} token of the two encoders, the Image Encoder and the Text Encoder, after passing them into a linear layer to project in a lower dimension space. Given an Image-Text pair $(I,T)$, we obtain $v_{cls}$  from $\mathcal{E}_v(I)$ and $t_{cls}$ from $\mathcal{E}_t(T)$. Then, the two embeddings are fed into the linear layer, obtaining $t'_{cls}$ and $v'_{cls}$. The same process is replicated also for the momentum model, obtaining  $\hat{t'}_{cls}$ and $\hat{v'}_{cls}$. Also, an image queue $\hat{Q}_i$ and a text queue $\hat{Q}_t$ are stored to implicitly enlarge the batch size. The CL is then formulated as:
\begin{equation}
    \mathcal{L}_{NCE}(x_1,x_2,Q)=-\frac{1}{|Q|}\sum_{(x,x_+)\in (x_1,x_2)}\log \frac{\exp(s(x,x_+)/\tau)}{\sum_{x_i\in Q}\exp(s(x,x_i)/\tau)}
\end{equation}
where $\tau$ is a learnable temperature parameters, $Q$ is the queue and $s(x,x_+)=\frac{x^Tx_+}{||x||\cdot||x_+||}$. The image-text constrative loss (ITC) \cite{li2021align,radford2021learning} is formulated as:
\begin{equation}
    \mathcal{L}_{ITC}=[\mathcal{L}_{NCE}(v'_{cls},\hat{t'}_{cls},\hat{Q}_t)+\mathcal{L}_{NCE}(t'_{cls},\hat{v'}_{cls},\hat{Q}_v)]/2
\end{equation}

\noindent
Other than $\mathcal{L}_{ITC}$, in RaSa also a intra-modal constrative loss (IMC) is added, which focuses on keeping close the image and text embedding of the same people with respect to the other people.

\begin{equation}
    \mathcal{L}_{IMC}=[\mathcal{L}_{NCE}(v'_{cls},\hat{v'}_{cls},\hat{Q}_v)+\mathcal{L}_{NCE}(t'_{cls},\hat{t'}_{cls},\hat{Q}_t)]/2
\end{equation}
\noindent
The final loss then becomes:
\begin{equation}
    \mathcal{L}_{CL}=(\mathcal{L}_{IMC}+\mathcal{L}_{ITC})/2
\end{equation}

\subsection{Attribute Loss}
Our attribute loss is designed to enhance the model capability to distinguish between matching and non-matching text-image pairs. In particular, we define an attribute in a caption as a chunk of words composed by a noun and its corresponding adjectives (\textit{e.g.} ``white long shirt"). To extract these chunks, SpaCy \cite{honnibal2020spacy} was employed. The idea behind this loss is that in captions composed by several attributes the model is not able to give the right importance to each attributes and potentially could ignore the most discriminative ones. Limiting this effect is crucial since often, due to the vague nature of text description, two people with different identities could be described by very similar texts, differing only for a single attribute. In this case, if most distinctive attributes are neglected, the correct matching between a text description and the correct person could fail, hindering the model accuracy. For this reason, the proposed attribute loss has the objective of limiting these cases, ultimately making the whole system more robust.


\begin{figure}[!th]
    \includegraphics[width=\textwidth]{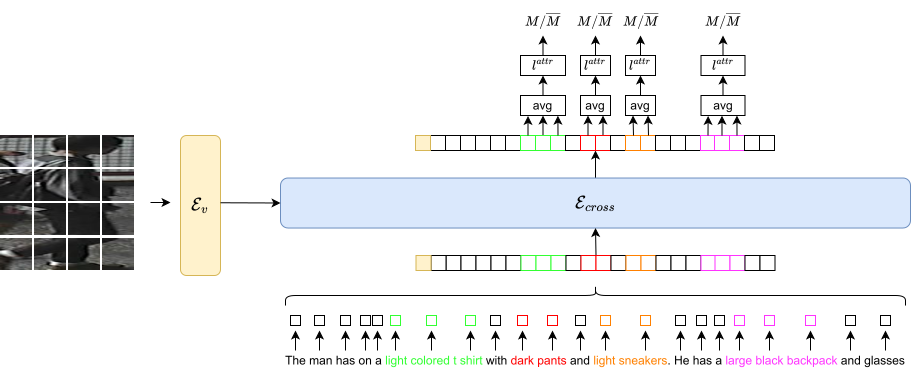}
    \caption{An overview of the Attribute Loss. 
    Using SpaCy, chunks of sentences containing nouns and related adjectives are identified. Then, after each token is processed by $\mathcal{E}_{cross}$, the average of each chunk embeddings is calculated. For each of them, the model then predicts if the image-chunk pair is a match or not. In the figure, chunks of words with the same color (\textit{i.e.} green, red, orange and purple) represent the extracted chunks and their corresponding embeddings (each box represents an embedding).}
    \label{fig:sal}
\end{figure}

In order to do so, given the output of the cross-modal encoder $\mathcal{E}_{cross}$, which takes as input the text $T$ and the image embedding $\mathbf{v}=\mathcal{E}_v(I)$, for each attribute \textit{i.e.} chunk \( ch \) of noun-adjective words in a given text $T$, the average of the corresponding embeddings is calculated as follows:


\begin{equation}
\hat{ch}(T,\mathbf{v},ch) = \frac{1}{N_w^{ch}}\sum_{w \in ch} \mathcal{E}_{cross}(T,\mathbf{v})[w^i]
\label{eq:chunk_avg}
\end{equation}
where $N_w^{ch}$ is the number of words in a given chunk $ch$ and $w^i$ is the position of the word $w$ in the output of $\mathcal{E}_{cross}$. 

Having this information, is now possible to calculate the proposed Attribute Loss $\mathcal{L}_\mathrm{AL}$ for each chunk. More in detail, $\mathcal{L}_\mathrm{AL}$ is tasked to perform a matching between each attribute chunk in the caption and the real image.  Let \( N_B \) be the batch size, \( N_{ch} \) the number of chunks in a text $T$ associated with an image $I$ and $l^{attr}_c(\hat{ch}(t,i,ch))$ be the same fully connected layer as the Eq. \ref{eq:lpitm} which predict if the image-text pair $(I, T)$ matches or not. The loss function becomes:

\begin{equation}
    \mathcal{L}_\mathrm{AL}=\frac{1}{3\cdot N_B}\sum_{(I,T)\in P}\frac{1}{N_{ch}}\sum_{ch\in t}\sum_{c\in C}y_c\log\frac{\exp(l^{attr}_c(\hat{ch}(T,\mathcal{E}_{v}(I),ch)))}{\sum_{n\in C}\exp(l^{attr}_n(\hat{ch}(T,\mathcal{E}_{v}(I),ch)))}
\end{equation}

\begin{figure}[!t]
    \centering
    \includegraphics[width=0.8\textwidth]{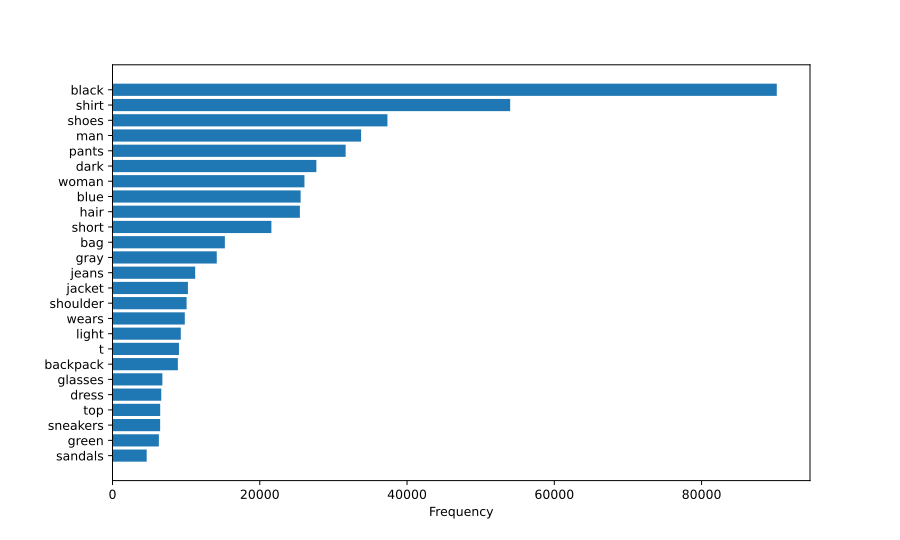}
    \caption{Top 25 most common nouns and adjectives in CUHK-PEDES computed using SpaCy \cite{honnibal2020spacy}}
    \label{fig:word_f}
\end{figure}

\noindent
Here, $C$, \( y_c \) and $P$ are built as in Eq. ~\ref{eq:lpitm}. 

Furthermore, we explored a weighted variant of the loss function. The results of this experiment are presented in Table ~\ref{tab:ablation} later in the paper. Specifically, we selected the top 25 most common nouns and adjectives in the CUHK-PEDES corpus (Fig. ~\ref{fig:word_f}) and calculated the frequency values normalized between 0 and 1. Let $\alpha_w$ denote the frequency of a given word $w$. If the word is not among the top 25 most common words, we set $\alpha_w$ to 0. We then define the importance weight $\omega_{ch}$ for the chunk $ch$ as follows:
\begin{equation}
    \omega_{ch}=1-\frac{\sum_{w\in ch}\alpha_w}{N^{ch}_w}
    \label{eq:weight}
\end{equation}
where $N^{ch}_w$ is the total number of words contained in the chunk. Finally, the final weighted attribute loss is formulated as:

\begin{equation}
    \mathcal{L}_\mathrm{weighted-AL}=\frac{1}{3\cdot N_B}\sum_{(I,T)\in P}\frac{1}{N_{ch}}\sum_{ch\in t}\sum_{c\in C}\omega_{ch}\cdot y_c\log\frac{\exp(l^{attr}_c(\hat{ch}(T,\mathcal{E}_{v}(I),ch)))}{\sum_{n\in C}\exp(l^{attr}_n(\hat{ch}(T,\mathcal{E}_{v}(I),ch)))}
\end{equation}
\noindent
As described in Eq. ~\ref{eq:weight}, lower importance weights ($\omega_{ch} \rightarrow 0$) are assigned to chunks with very common words and higher importance weights  ($\omega_{ch} \rightarrow 1$) are assigned to chunks with uncommon words. This approach is used to downweigh the contribution of very common attributes that match with several different images and therefore identities.

In summary, attribute loss is used to pay attention on the subtle details of a single sentence, improving matching performance using fine-grained details contained in the text that describe an image (\textit{i.e.} ``A pink headset'' can be a very uncommon attribute that, if properly considered, improves the model accuracy). As a result, attribute loss helps the model to use the entire given text without losing details. In other words, by distributing the attention evenly, it encourages a more comprehensive understanding of the input data.

\subsection{Masked AutoEncoder Loss}
Inspired by the masked language model, we have developed a novel loss function based on the Masked AutoEncoder \cite{he2022masked} (MAE). MAE was originally used as a self-supervised training technique for transformers. The goal is to reconstruct a sequence of masked image patches back into the original unmasked one. In our case, we customized this technique integrating also text embeddings. More in detail,
we inject the text embeddings in the MAE decoder via cross attention layers. The aim is to use the textual information to help the decoder reconstruct the image patches, hence strongly linking together words and visual information.

Given an image-text pair $(I,T)$, we randomly sample patches from the image $I$ with a probability $p_{mae}$ and discard the remaining patches. The selected patches are processed through the Image Encoder $\mathcal{E}_v$ to obtain their corresponding embeddings $\{v_{cls},v_1,\dots,v_L\}$, with $L<M$. Prior to feeding these embeddings into the MAE decoder $\mathcal{D}_{mae}$, the embeddings for the removed $K=M-L$ patches are replaced with a learnable mask embedding, thus obtaining a set $\mathbf{v}_{masked}=\{v'_{cls},v'_1,\dots,v'_{M}\}$ of dimension $M$\@. The set $\mathbf{v}_{masked}$ is then fed into the MAE decoder $\mathcal{D}_{mae}$, where it is fused with the text embeddings  $\{t_{cls}, t_1,\dots,t_N\}=\mathcal{E}_t(T)$ corresponding to the text $T$ using cross-attention mechanism to reconstruct the original image. The MAE loss is a reconstruction loss, which is calculated using the mean squared error (MSE) of the removed patches:

\begin{equation}
    \mathcal{L}_\mathrm{MAE}=\frac{1}{N_B}\sum_{i=0}^{N_B}\frac{1}{K}\sum_{j=0}^{M}m_i^j||x_j^i-\hat{x}_j^i||_2^2
\end{equation}
\noindent
where \( m_i^j \) is an indicator variable that equals \( 1 \) if the patch was originally removed and thus needs to be reconstructed, and \( 0 \) otherwise. Let $x_j^i$ be the original image patch and $\hat{x}_j^i$ be the reconstructed one, then, $\hat{x}_j^i=\mathcal{D}_{mae}(\mathbf{v}_{masked},\mathcal{E}_t(T))$.

In our case the proposed MAE is trained end-to-end along with all the other components of the model bridging the gap between textual and image information.

\subsection{Full Objective and Reranking}
Finally, the complete model loss is:
\begin{equation}
    \mathcal{L}=\underbrace{\mathcal{L}_{p-ITM}+\lambda_1\mathcal{L}_{prd}}_{\mathcal{L}_{RA}}+\underbrace{\mathcal{L}_{MLM}+\lambda_2\mathcal{L}_{m-RTD}}_{\mathcal{L}_{SA}}+\lambda_3\mathcal{L}_{CL}+\lambda_4\mathcal{L}_{MAE}+\lambda_5\mathcal{L}_{AL}
    \label{eq:final_loss}
\end{equation}
\noindent where each $\lambda_*$ is a weight assigned to a specific loss.

During inference, referring to both ALBEF \cite{li2021align} and RaSa \cite{bai2023rasa}, considering the high inefficiency of the quadratic interaction operation, we employ a sampling strategy, where we select a subset of k image-text pairs and apply the ITM rank to this reduced set. Specifically, given a text input $T$, we identify the top-k, with $k=128$, images by computing the similarity scores $s(t_{cls},v_{cls})$ and selecting the images with the highest scores. An analysis of how changing this parameter affects both efficiency and accuracy is provided in Section ~\ref{sec:top-k}.

\section{Experimental Results}

\subsection{Experimental Settings}
We train our model on a single NVIDIA 4090 GPU for a total of 30 epochs using a batch size of 8. We employ the AdamW optimizer \cite{loshchilov2018decoupled} with a weight decay of $0.02$ decay. Initial values of the learning rate are $1e-4$ for PRD and $m$-RTD parameters, and $1e-5$ for other parameters. Images are resized to $384\times 384$ (dataset image size is $128\times 384$), with also the possibility of horizontal random flip. We set the maximum number of words in BERT to 70. Momentum coefficient $m$ is set to $0.995$. The temperature $t$ is set to $0.07$, and the queue size utilized in the CL loss is $65536$. With regard to the mask ratio, we have it set at 75\%, thus 75\% of image patches are eliminated before going through $\mathcal{E}_v$. We employ the standard BERT \cite{devlin2018bert} for the $MLM$ loss, with a masking probability of 15\%, while, for the $m-RTD$ loss, a masking probability of 30\% is used. Finally, the probability of inputting a weak pair in RA is set to $0.1$. We set the $\lambda$s of the loss described in Eq ~\ref{eq:final_loss} as $\lambda_1=0.5,\:\lambda_2=0.5,\:\lambda_3=0.5,\lambda_4=1,\:\lambda_5=2$.

\subsection{Metrics}
To evaluate our model, we adopt widely-used metrics in TBPS. Firstly, we evaluate our model with Rank@K, with K=1, 5 and 10. Rank@K evaluates how many times a model is able to predict at least an image corresponding to a given text in the first K proposed images. Lastly, we calculate the mean Average Precision (mAP). Let $N_T$ be the number of text in the test set, we calculate the mAP as the mean of each average precision for each text $t$ ($AP_t$).

\begin{equation}
    mAP=\frac{1}{N_{T}}\sum_{t\in T}AP_t
\end{equation}
AP expresses how well the model is able to retrieve correct images in the early positions. It can be calculated as:
\begin{equation}
    AP=\frac{1}{N_{id}}\sum_{k} P(k)\cdot \mathrm{rel}(k)
\end{equation}
where $N_{id}$ is the number of the correct identities, $P(k)$ is the precision at the position $k$, calculated as $\frac{\sum_{i=1}^k m_i}{k}$, with $m_i=1$ if it is a correct match, 0 otherwise and $\mathrm{rel}(k)$ is the indicator function which is 1 if the position $k$ contains a positive match, $0$ otherwise.

We argue that mAP is a crucial metric to express the quality of a retrieval model since it encapsulates better the capability of the model to propose positive match in top positions. This is especially true for TBPS where we want to be able to find all the different identities corresponding to a specific caption.

\subsection{Datasets}
We trained and tested our model on three different standard datasets.
\begin{itemize}
    \item CUHK-PEDES \cite{li2017person}: composed by 40206 images of pedestrians with 13003 different identities. Each image is paired with 2 text descriptions. The first one contains a coarse description of the image, while the second one is more fine-grained and rich in details. Among all the different identities, 1000 are used for the evaluation phase.
    \item ICFG-PEDES \cite{ding2021semantically}: containing 54522 pedestrian images divided into 4102 unique identities. The text information is more fine-grained and identity-centric than CUHK-PEDES. It is divided into a training and a testing set having 34674/19848 images and 3102/1000 identities, respectively.
    \item RSTPReid \cite{zhu2021dssl}: constructed with 25505 images having 4101 different identities. 15 cameras were used to collect the dataset and each person is represented by 5 images in the dataset each having 2 textual descriptions. The dataset is divided in training, validation and testing set having 3701, 200 and 200 identities, respectively.
    
\end{itemize}

\begin{table}[]
    \centering
    
    \adjustbox{max width=\textwidth}{%
    \begin{tabular}{c|cccc|cccc|cccc}
    \hline 
     & \multicolumn{4}{c|}{CUHK-PEDES}& \multicolumn{4}{c|}{ICFG-PEDES} & \multicolumn{4}{c}{RSTPReid} \\
    \hline
    Model & R@1 & R@5 & R@10 & mAP & R@1 & R@5 & R@10 & mAP & R@1 & R@5 & R@10 & mAP\\
    \hline
    ALBEF \cite{li2021align} & 60.28 & 79.52 & 86.34 & 56.67 & 34.46 & 52.32 & 60.40 & 19.62 & 50.10 & 73.70 & 82.10 & 41.73 \\
    BLIP \cite{li2022blip} & 64.36 & 83.36 & 88.78 & 58.18 & 56.16 & 73.77 & 80.17 & 31.59 & - & - & - & -\\
    CLIP (VIT-B/16) \cite{radford2021learning} & 68.19 & 86.47 & 91.47 & 61.12 & 56.74 & 75.72 & 82.26 & 31.84 & 54.05 & 80.70 & 88.00 & 43.41 \\
    \hline
    CFine \cite{yan2023CLIPdrive} & 69.57 & 85.93 & 91.15 & - & 60.83 & 76.55 & 82.42 & - & 50.55 & 72.50 & 81.60 & - \\
    IRRA \cite{jiang2023cross} & 73.38 & 89.93 & 93.71 & 66.13 & 63.46 & 80.25 & \textbf{85.82} & 38.06 & 60.20 & 81.30 & 88.20 & 47.17 \\
    TBPS-CLIP \cite{cao2024empirical} & 73.54 & 88.19 & 92.35 & 65.38 & 65.05 & 80.34 & 85.47 & 39.83 & 61.95 & 83.55 & 88.75 & 48.26 \\
    CADA$^*$ \cite{lin2024cross} & \underline{77.20} & \textbf{90.68} & 93.92 & 68.45 & \underline{67.38} & \underline{81.34} & 85.64 & 37.81 & \textbf{67.70} & 84.60 & \underline{89.75} & 49.95 \\
    RaSa \cite{bai2023rasa} & 76.51 & 90.29 & \underline{94.25} & \underline{69.38} & 65.28 & 80.40 & 85.12 & \underline{41.29} & 66.90 & \underline{86.50} & \textbf{91.35} & \underline{52.31} \\
    \hline
    \textbf{MARS (Ours)}     & \textbf{77.62} & \underline{90.63} & \textbf{94.27} & \textbf{71.41} & \textbf{67.60} & \textbf{81.47} & \underline{85.79} & \textbf{44.93} & \underline{67. 55} & \textbf{86.65} & \textbf{91.35} & \textbf{52.92} 
    \end{tabular}
    }
    
    \caption{Results of state-of-the-art models compared with Ours on CUHK-PEDES, ICFG-PEDES and RSTPReid. * model retrained since no checkpoints were available} 
    \label{tab:results}
\end{table}
\subsection{Results Analysis}
Table \ref{tab:results} presents a comprehensive comparison of the proposed model with state-of-the-art models on three benchmark datasets: CUHK-PEDES, ICFG-PEDES, and RSTPReid. The results demonstrate the effectiveness of the proposed model in terms of Rank@1 (R@1), Rank@5 (R@5), Rank@10 (R@10), and mean Average Precision (mAP). 

First of all, in the first three lines of the table are presented the results obtained by directly finetuning three pretrained large vision-language models such as CLIP, BLIP and ALBEF. Then, a set of the best current state-of-art models is presented. To ensure a fair comparison, since our model was pretrained on ALBEF, they all belongs to the family of TBPS models pretrained on the aforementioned large-language models. Finally the results of the proposed system are presented.

On the CUHK-PEDES dataset, our model outperforms all other SOTA models, achieving the highest performance on all the proposed metrics except for R@5, where our model is still the second best. Specifically, the proposed model surpasses the previous SOTA models, by $0.42$ in R@1, $0.02$ in R@10, and a significant $2.03$ in mAP. 

On the ICFG-PEDES dataset,  the proposed model manages to surpass all the other SOTA models, except for R@10, where our model is still the second best. More importantly, our model obtains the highest mAP score by $3.64$. This proves that our model works better when the captions are more fine grained and identity-centric like the ones of ICFG-PEDES. Indeed, in this case for the attribute loss is easier to boost the contribution of each attribute chunk in the textual descriptions. 

On the RSTPReid dataset, the proposed model achieves a slightly lower R@1 score compared to CADA ($0.15$ less), but outperforms other SOTA models in R@5, R@10, and mAP. Since RSTPReid is a very small dataset compared to the others, this demonstrate the ability of our model to be more robust to overfitting that previous methods. 

Overall, our model was able to improve the results on all three most common TBPS dataset, demonstrating its robustness to diverse and challenging scenarios.

The proposed model strength lies in its ability to accurately rank relevant results. This is proven by the fact that the proposed model mAP is consistently higher than the other models. Indeed, a higher mAP allows the system to accurately retrieve all the identities corresponding to a specific caption in the initial ranking position which is crucial for TBPS.

\section{Ablation}
We perform our ablation experiments on CUHK-PEDES and then train the best model on the other two datasets.

\begin{figure}[h]
    \centering
    \includegraphics[width=\textwidth]{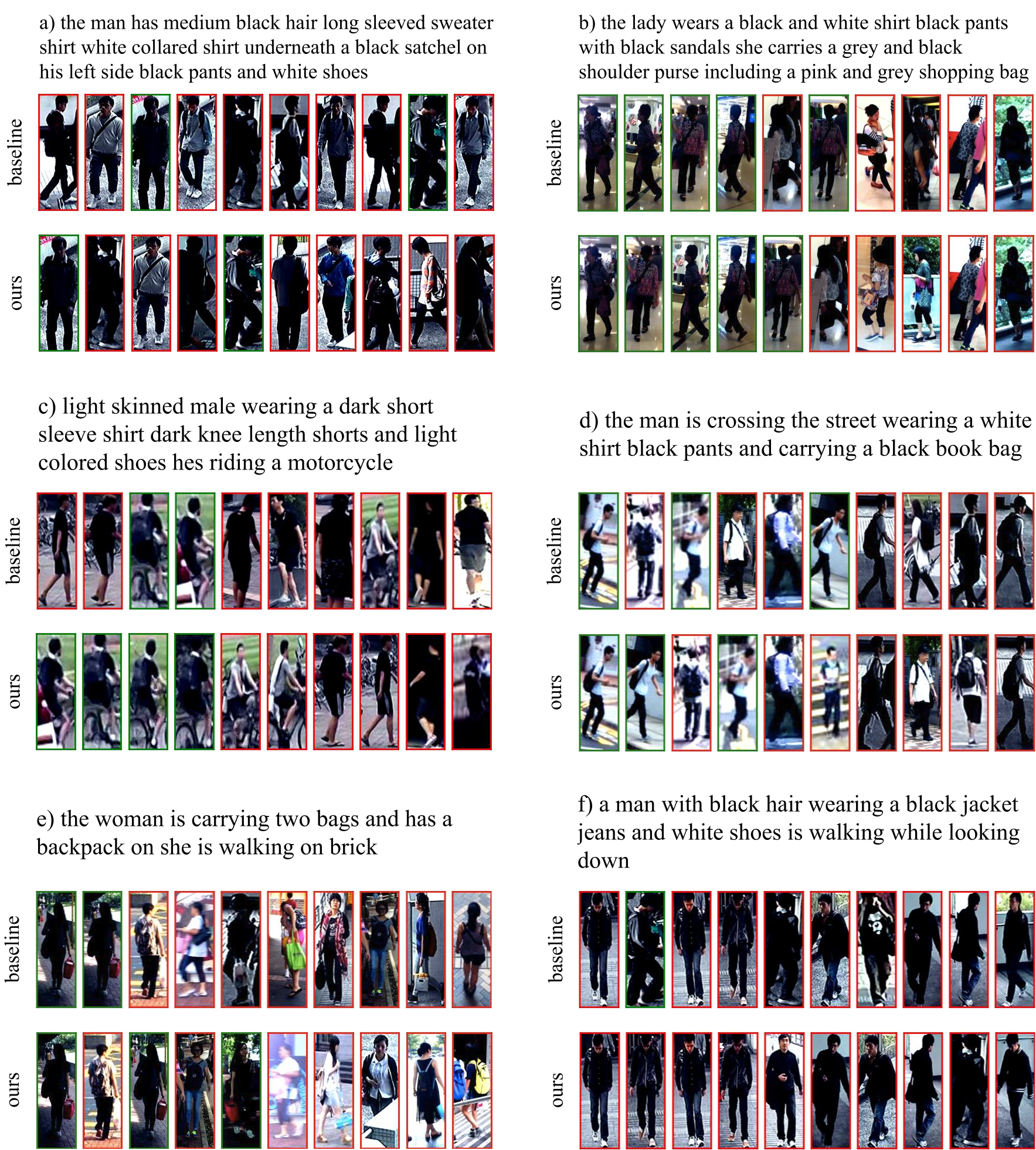}
    \caption{Overview of comparison between top 10 predictions of baseline and our model. Predicted images are ranked from left (\textit{i.e.}, position 1) to the right (\textit{i.e.}, position 10). Our model outperforms the baseline in several pairs, \textit{i.e.}, \textit{a,b,c,d}. In pair \textit{c} it is possible to observe how all predictions are with a bike in it, while this is not true in the baseline. Furthermore, even if in pair \textit{e} our model does not predict the second position correctly, it is easy to observe how a higher mAP is achieve by providing 3 correct matches in top 10 positions compared to 2 correct matches in top 10 of the baseline. Lastly, in pair \textit{f} our model is not able to predict any correct image due to the vagueness of the caption, but is still retrieving images closely related to the text.}
    \label{fig:topk}
\end{figure}

\begin{figure}[h]
    \centering
    \includegraphics[width=0.8\textwidth]{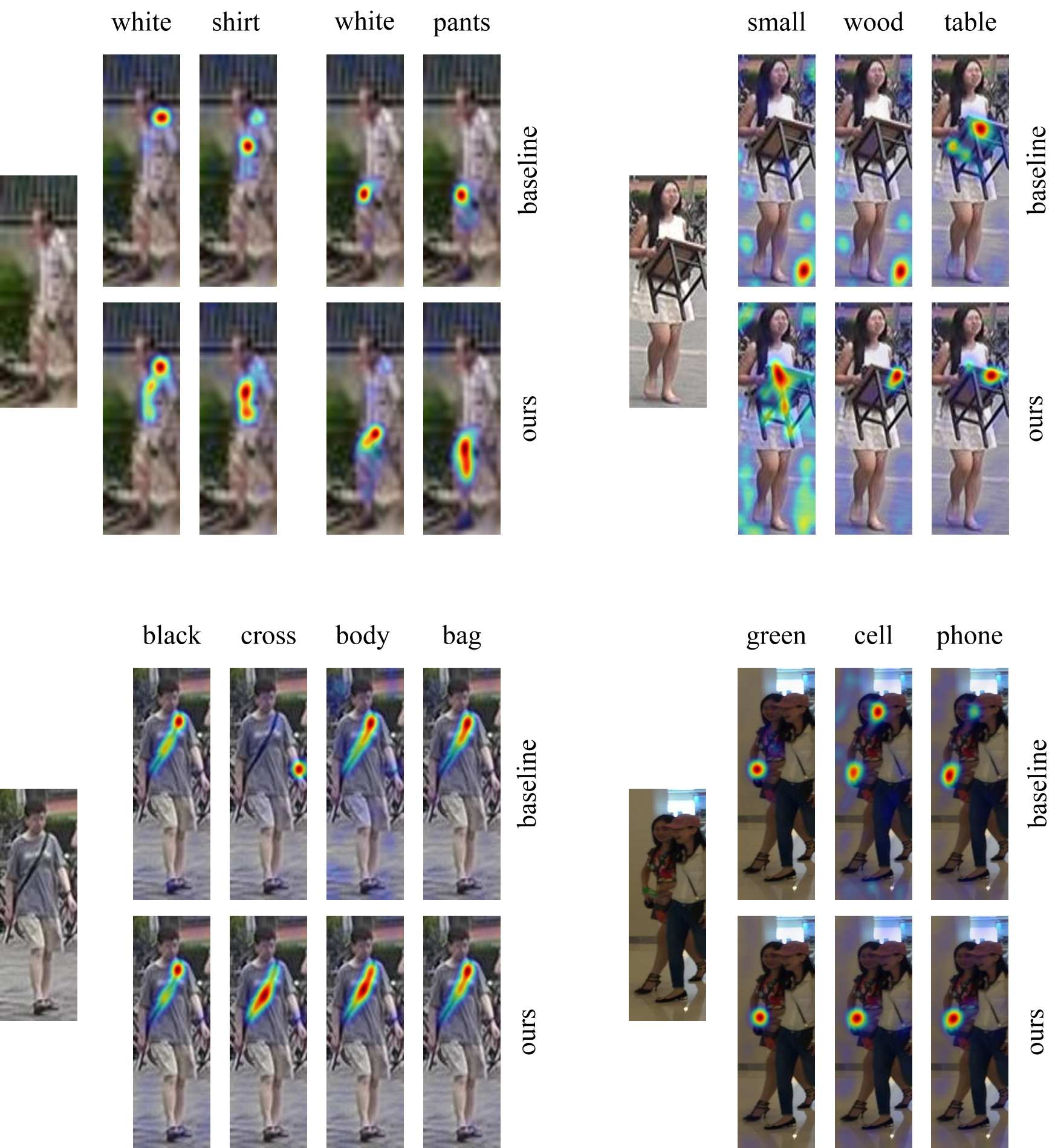}
    \caption{Visual comparison of cross attention maps generated by the baseline model (top) and our model (bottom) using Grad-CAM \cite{selvaraju2017grad}. The attention maps illustrate the cross-modal encoder focus on different regions corresponding to individual words in the attribute chunks. The proposed attribute loss leads to more consistent and accurate attention distribution across words.}
    \label{fig:gradcam}
\end{figure}

Starting from a qualitative ablation, in Fig. ~\ref{fig:topk} it is possible to observe the difference in ranking between the baseline model (RaSa) and our model. In each image-text pair, the images are the top 10 ordered from left to right, where left is the one with highest probability of matching. In these examples, a better R@1 and mAP can be seen in image-text pair \textit{a}, where ours model is able to predict the correct image in the first position and also provide another correct image in a higher position compared to the baseline model, the same happens with pair \textit{d}. In pair \textit{b} our model classifies all the first top 5 images correctly. Additionally, in pair \textit{c} our model focuses more on all attributes contained in the given text. Indeed, beside the fact that the first top 4 images are all correct it is possible to observe how also in position 5 and 6 there is a person riding a bike while this behavior is not observable in the baseline. Furthermore, in pair \textit{e} our model is able to achieve higher mAP, even if, compared to the baseline, our second prediction is wrong. Indeed, we are able to predict 3 correct images in the top 10 when the baseline only predicts two correct ones in its top 10. Lastly, we provide in the pair \textit{f} an example of failure of our model, where it is not able to predict any correct image in top 10. However, it is worth noting that all the predicted images of our model are very related to the given caption. In this case this failure is probably due to the intrinsic vagueness of the text captions that often are very difficult to be linked to a specific identity.

Moreover, in Fig \ref{fig:gradcam} several visual comparisons between the baseline model and the model trained with the proposed attribute loss are presented. More in detail, Grad-CAM algorithm \cite{selvaraju2017grad} was employed to extract attention maps in the cross-modal encoder, each corresponding to the attention of a single word w.r.t the whole person image. In the figure, some attribute chunks were chosen to highlight the effect of the attribute loss. In particular, it can be appreciated how the attention produced using our model is much more consistent over all the words that compose the attribute chuck. This is particularly evident in the "black cross-body bag" and "small wood table" attributes, where the attention is distributed over the correct object for each of the words. In addition, our system allows to generate attention maps that focus much more over the correct attribute. For example, in the attributes "white shirt" and "white pants" the attention maps of our architecture are spread over the corresponding clothes more uniformly than the baseline architecture. On the other side, for the attribute "green cell-phone" no undesired attention noise can be found with our model, whereas the baseline focus is also on random parts of the image. Indeed, these qualitative results help to validate our approach by demonstrating that the proposed training objective helps to precisely link text and image information, which is crucial for the TBPS task.

\begin{table}[t!]
    \centering

    \begin{tabular}{c|ccc|cccc}
    \hline
    Model & MAE & AL & Full CA & R@1 & R@5 & R@10 & mAP \\
    \hline
    Baseline (RaSa) & \xmark{} & \xmark{} & \xmark{} & 76.51 & 90.29 & \underline{94.25} & 69.38 \\
    Baseline with 70 words & \xmark{} & \xmark{} & \xmark{} & 77.03 & 90.24 & 94.15 & 70.03 \\
    \hline 
    A1 & \vcheck{} & \xmark{}  & \xmark{}  & 77.08 & 90.07 & 94.18 & 69.88 \\
    A2 & \xmark{}  & \vcheck{} & \xmark{}  & 76.92 & 90.24 & 94.01 & 70.92\\
    A3 & \xmark{}  & \xmark{}  & \vcheck{} & 77.18 & 89.90 & 93.44 & 70.48 \\
    A4 & \vcheck{} & \vcheck{} & \xmark{}  & \underline{77.63} & 90.22 & 94.10 & 71.35\\
    A5 & \vcheck{} & \xmark{}  & \vcheck{} & 77.06 & 90.11 & 93.92 & 70.07 \\
    A6 & \xmark{}  & \vcheck{} & \vcheck{} & 77.45 & \underline{90.48} & 94.22 & \textbf{71.46} \\
    \hline
    Ours w/o shared head     & \vcheck{} & \vcheck{} & \vcheck{} &  77.18 & 90.16 & 93.89 & 71.20 \\
    Ours with AL rebalanced & \vcheck{} & \vcheck{} & \vcheck{} & \textbf{77.84} & 90.27 & 94.04 & 71.19 \\
    \hline
    \textbf{MARS (Ours)}     & \vcheck{} & \vcheck{} & \vcheck{} &  77.62 & \textbf{90.63} & \textbf{94.27} & \underline{71.41} \\
    \end{tabular}

    \caption{Ablation study performed on CUHK-PEDES. First two rows represent the baseline (\textit{i.e.}, RaSa) and the baseline trained with caption capped at 70 words instead of 50 words. Other rows (from A1 to A6) show the results of our model with all possible combination of losses (MAE loss, Attribute Loss and Full CA that is the text encoder with additional cross attention). Additional ablations with all the losses are provided. The former is the model trained without using the same head for Attribute Loss and Relation-Aware Loss. The latter is the model trained using the rebalanced version of our Attribute Loss.}
    \label{tab:ablation}

\end{table}
Finally, in Table \ref{tab:ablation} a quantitative comparisons between all the different possible configurations of our model is presented. More in detail, the first two rows report two different RaSa \cite{bai2023rasa} baseline versions with the second one which consider a maximum sentence length of 70 instead of 50 for RaSa. Since the incremented sentence length proved to be better, we have chosen to employ that configuration for all the following experiments. In particular, tests from A1 to A6 represent all the different combinations of training our model with the masked autoencoder loss, the attribute loss and cross-attention layers in each of the 12 blocks of the Cross Modal Encoder. We decided to comment these results focusing our analysis on the attribute loss and the effect that the other losses have on it. 
Surprisingly, this loss alone (test A2) is not able to boost the R@1 score of the model (76.92 vs 77.03), but the mAP is increased (70.92 vs 70.03), meaning that more correct images are found earlier in the retrieval rank. On the contrary, when paired with the masked autoencoder loss (test A4) or the increased cross-attention layers (test A6), the attribute loss is able to improve the overall performance of the model. The motivation for this is twofold. On one side, the MAE loss is able to increase the connection between single words and image patches which benefits also the attribute loss. On the other side, more cross-attention layers means a better interaction between image and text embeddings. Indeed, the importance of the attribute loss is confirmed by the fact that, when the model is trained without it (tests A1, A3 and A5) the quantitative results do not improve w.r.t. the baseline. 

Finally, the last two ablations consist in training the attribute loss using a different head than the one used to perform the global matching and re-balancing the attribute loss with weights calculated considering the frequency of the words in the dataset. Quantitative results confirm that sharing the matching head also to perform attribute matching is beneficial to the model and therefore we decided to use this as a final configuration. On the other side, a weighted attribute loss allows to achieve an higher R@1, but it performs worse overall in the other metric and therefore we choose not to use it in our final configuration.

\subsection{Effect of changing top k for ITM ranking}\label{sec:top-k}
\begin{figure}[h]
    \centering
    \includegraphics[width=\textwidth]{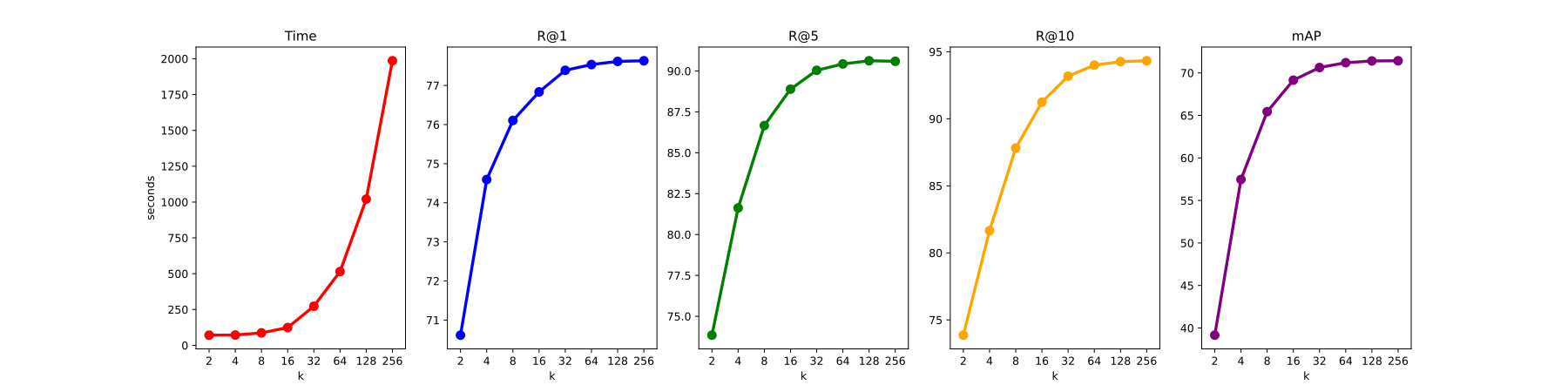}
    \caption{Impact of varying the sampling parameter k for the ITM ranking on the performance of the model, tested on CUHK-PEDES. The plots show the trade-off between computational efficiency (first plot) expressed in seconds to evaluate the entire test set and the accuracy, expressed as R@1, R@5, R@10 and mAP.
    We set $k$ equals to 128, since, even if higher values allows the model to obtain better accuracy, these improvements are not justified by the additional required evaluation time.}
    \label{fig:eval-k}
\end{figure}

We select $k=128$ for ITM ranking by exploring the effect of changing it on several measures, such as time to perform the re-ranking, R@1, R@5, R@10 and mAP. As it can be observed in Fig. \ref{fig:eval-k} increasing $k$ is beneficial for the other metrics, except execution time. In addition, after $k=128$ the positive effect becomes almost negligible. Indeed, the larger requested time for $k=256$, which is almost the double w.r.t. $k=128$, does not justify the accuracy gain, which is only $0.016$, considering R@1 only.

\subsection{Efficacy of Attribute Loss}

\begin{figure}[h]
    \centering
    \includegraphics[width=\textwidth]{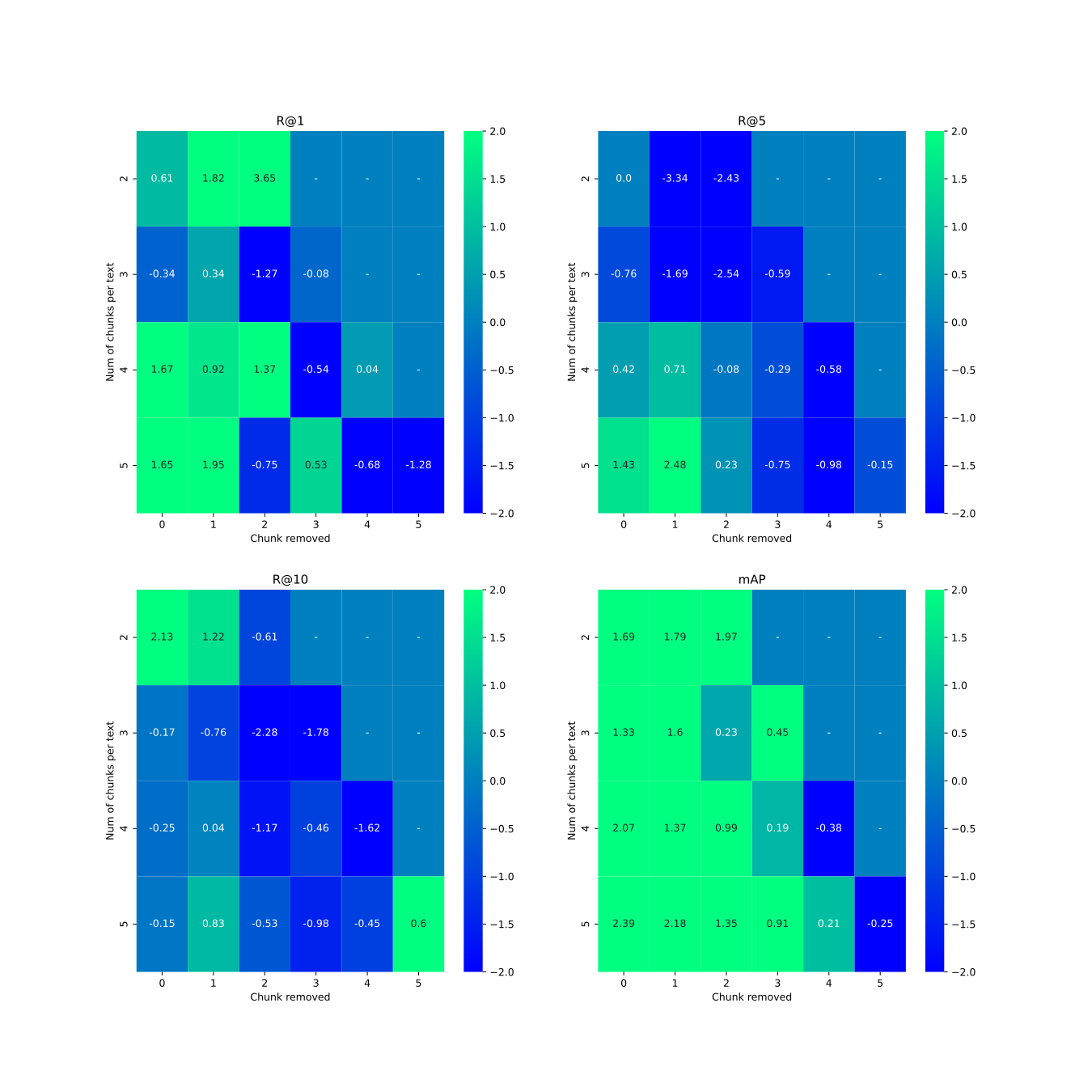}
    \caption{Results when removing attribute chunks from captions. Each cell contains the difference between the results of our model and the baseline model. Green indicates that our model performs better than the baseline while blue indicates the opposite.}
    \label{fig:heatmap}
\end{figure}

Finally, we want to demonstrate the efficacy of the proposed attribute loss to boost the importance of each attribute chunk in a caption in order to be equally considered w.r.t. the others during the searching process. In order to prove that, we designed the following experiment: firstly, sentences are divided based on the number of attribute chunks (from 2 to 5); then, different number of chunks (from 0 to 5) are randomly removed from the sentences and the resulting value of R@1, R@5, R@10, and mAP are calculated. 

We performed this test both on the baseline model and the proposed model and results can be seen in Fig.~\ref{fig:heatmap}. More in detail, each cell in the figure represents the difference between our model and the baseline of the corresponding metric value, therefore green colored cells indicate that our model was more resilient than the baseline to the attribute chunk removal, while blue colored cells indicates the opposite.

As expected, thanks to the attribute loss, our model was able to perform much better than the baseline even after the attribute chunk removal. This is especially true for sentences with a higher number of attribute chunks (4 or 5). Indeed, in these cases, removing a single attribute that is crucial in the caption could cause a catastrophic drop in performance in the baseline, while, since in our model each attribute is considered with the same importance, this effect is strongly mitigated. Notably, the mAP metric is basically always better for our model than the baseline. This means that, for each caption, we are always able to retrieve a higher number of the correct identities in better ranking positions. By considering all the attributes equally, some images that were neglected by previous models are now correctly considered during the search. 

It is also worth noting that our model performs worse than the baseline especially in cases where a higher number of chunks is removed. In this case, it is important to consider the fact that, when removing a high number of chunks, the accuracy in the retrieval drops dramatically both in the baseline and in our model making it difficult to perform an accurate analysis of the results. At the same time, this indicates that our model performs a search heavily based on attributes and is not able to perform well if no attribute chunk is found in the caption.

\section{Conclusions}
In this paper we proposed a novel architecture for TBPS named MARS which is composed by a text encoder, an image encoder and a cross-modal encoder, like some of the previous state-of-the-art systems, but, in addition, is also equipped with a masked autoencoder sharing the encoder part with the image encoder and implementing a decoder that takes masked image embeddings as input as well as textual embeddings.

Our proposed MARS architecture brings along a significant improvement in text-based person search. We develop a novel way to address the inter-identity and intra-identity variation, providing a robust solution which is capable to outperform the current state of the art.

Specifically, thanks to the masked autoencoder, we develop a new visual reconstruction loss, which manages to encourage the model to learn a more informative embedding coming from both text and image encoder. Secondly, we equip the whole cross-modal encoder with additional cross attention for the reranking phase. Lastly, we develop a novel attribute loss, which enables the model to focus on every attribute of a given sentence. It is worth noting that, as shown by our ablation this loss alone is not able to push the model to its best, but when coupled with MAE Loss or the new cross model encoder, the attribute loss allows the model to outperform the state of the art.

As a conclusion, all the aforementioned novelties make MARS a model with outstanding performances, especially w.r.t. the mAP. This means that overall, our model is able to rank matching results in earlier positions than previous methods which is crucial in a real world scenario.

\begin{acks}
This work was funded under the National Recovery and Resilience Plan (NRRP), Mission 4 Component 2 Investment 1.4 - Call for tender No. 3138 of 16/12/2021 of Italian Ministry of University and Research funded by the European Union – NextGenerationEU, Project code CN00000023, Concession Decree No. 1033 of 17/06/2022 adopted by the Italian Ministry of University and Research, CUP D93C22000400001, “Sustainable Mobility Center” (CNMS).

Additionally, this work was partially supported by ``Partenariato FAIR (Future Artificial Intelligence Research) - PE00000013, CUP J33C22002830006" funded by the European Union - NextGenerationEU through the italian MUR within NRRP.
\end{acks}

\bibliographystyle{ACM-Reference-Format}
\bibliography{main}




\end{document}